\documentclass[runningheads]{llncs}
\usepackage{multirow}
\usepackage{array}
\usepackage{booktabs}
\usepackage[colorlinks=true, linkcolor=blue, citecolor=green, urlcolor=cyan]{hyperref}
\usepackage{pdfpages}

\usepackage[T1]{fontenc}
\usepackage{graphicx,verbatim}
\begin{document}
\title{ColorGS: High-fidelity Surgical Scene Reconstruction with Colored Gaussian Splatting}
\titlerunning{ColorGS for High-Fidelity Surgical Scene Reconstruction}

\author{Qun Ji, Peng Li, Mingqiang Wei}  
\institute{Nanjing University of Aeronautics and Astronautics \\
    \email{}}

\maketitle              
\begin{abstract}
High-fidelity reconstruction of deformable tissues from endoscopic videos remains challenging due to the limitations of existing methods in capturing subtle color variations and modeling global deformations. While 3D Gaussian Splatting (3DGS) enables efficient dynamic reconstruction, its fixed per-Gaussian color assignment struggles with intricate textures, and linear deformation modeling fails to model consistent global deformation. To address these issues, we propose ColorGS, a novel framework that integrates spatially adaptive color encoding and enhanced deformation modeling for surgical scene reconstruction. First, we introduce Colored Gaussian Primitives, which employ dynamic anchors with learnable color parameters to adaptively encode spatially varying textures, significantly improving color expressiveness under complex lighting and tissue similarity. Second, we design an Enhanced Deformation Model (EDM) that combines time-aware Gaussian basis functions with learnable time-independent deformations, enabling precise capture of both localized tissue deformations and global motion consistency caused by surgical interactions. Extensive experiments on DaVinci robotic surgery videos and benchmark datasets (EndoNeRF, StereoMIS) demonstrate that ColorGS achieves state-of-the-art performance, attaining a PSNR of 39.85 (1.5 higher than prior 3DGS-based methods) and superior SSIM (97.25\%) while maintaining real-time rendering efficiency. Our work advances surgical scene reconstruction by balancing high fidelity with computational practicality, critical for intraoperative guidance and AR/VR applications.

\end{abstract}
\section{Introduction}

Reconstructing 3D surgical scenes from endoscopic videos is essential for intraoperative navigation~\cite{ref_article9}, enhanced visualization~\cite{ref_article10,ref_article11}, and robotic-assisted surgery (RAMIS)~\cite{ref_article1}. Beyond providing real-time guidance, it also supports AR/VR-based surgical training~\cite{ref_article2} and preoperative planning by providing high-fidelity anatomical models. 
Moreover, improved 3D reconstruction contributes to optimizing surgical workflows, facilitating remote monitoring~\cite{ref_article4}, immersive visualization, and accelerated skill acquisition~\cite{ref_article3}. 
Thus, achieving accurate and efficient reconstruction is indispensable for advancing surgical practices and improving patient care.

Early methods for surgical scene reconstruction mainly rely on depth estimation~\cite{ref_article5} and SLAM-based point cloud fusion ~\cite{ref_article6,ref_article1}. However, these methods often suffer from inefficiencies due to complex processing pipelines and redundant data integration, limiting both real-time performance and accuracy. 
Recent studies, such as EndoNeRF~\cite{EndoNeRF} and EndoSurf~\cite{EndoSurf}, have shifted towards using neural radiance fields (NeRF)~\cite{ref_article8}, combined with differentiable rendering techniques to achieve high-quality and efficient surgical scene reconstruction.
Despite the improved reconstruction quality and streamlined pipeline, the high computational cost of NeRF leads to long training times and slow rendering speeds, significantly hindering its applicability in real-time surgical scenarios.

The emergence of 3D Gaussian Splatting (3DGS)~\cite{3DGS} marks a significant breakthrough, enabling high-fidelity reconstruction with greater efficiency compared to NeRF.
It has been rapidly extended to dynamic scene reconstruction tasks through the integration of deformation fields \cite{DBLP:conf/cvpr/WuYFX0000W24} and further introduced for dynamic medical scene reconstruction.
%
%
For example, EndoGaussian~\cite{EndoGaussian} models the deformation field with two lightweight modules and employs a multi-resolution HexPlane~\cite{Hexplane} as the 4D structural encoder. 
Building on ~\cite{Gaussian-flow,spacetime}, Deform3DGS~\cite{Deform3DGS} discards the time-consuming MLP-based deformation fields in favor of basis functions to model Gaussian motion, further enhancing the efficiency of 3D reconstruction. 
Despite the improved rendering speed for real-time deformational medical applications, they still struggle to achieve accurate modeling of realistic colors and global deformations due to the following limitations: 
%
\textbf{1) Insufficient Color Expressiveness.} In surgical scenes, tissues often exhibit similar appearances with subtle color variations. However, existing methods typically assign fixed color attributes to each Gaussian, leading to indistinguishable rendering results for similar tissues.
%
\textbf{2)  Locality of Gaussian Functions.} Due to the computational inefficiency of MLP-based deformation fields, existing methods exploit linear combinations of basic functions with trainable parameters to model motion efficiently. However, the localized nature of Gaussian functions \cite{Deform3DGS} makes it challenging to capture consistent global motion trends.

To this end, we propose ColorGS, a novel framework designed to enhance both color expressiveness and deformation modeling in 3DGS-based high-fidelity surgical scene reconstruction.
%
Firstly, we introduce spatially adaptive color modeling (Colored Gaussians), which enhances the color expressiveness of individual Gaussians by integrating color information from dynamic anchors tailored for each Gaussian primitive.
Furthermore, we propose a simple yet effective module, the Enhanced Deformation Model (EDM), which introduces a time-independent global deformation parameter to model consistent global motion and enhance the overall consistency and smoothness of the deformation.
By jointly improving color expressiveness and deformation accuracy, ColorGS effectively enhances both the quality and reliability of surgical scene reconstruction. 
Extensive experiments on the EndoNeRF~\cite{EndoNeRF} and StereoMIS~\cite{StereoMIS} datasets demonstrate that the proposed method achieves state-of-the-art reconstruction quality (PSNR: 39.85, SSIM: 97.25, LPIPS: 0.03).
%
%

\section{Method}
\subsubsection{Pipeline.} 
\begin{figure}
\includegraphics[width=\textwidth]{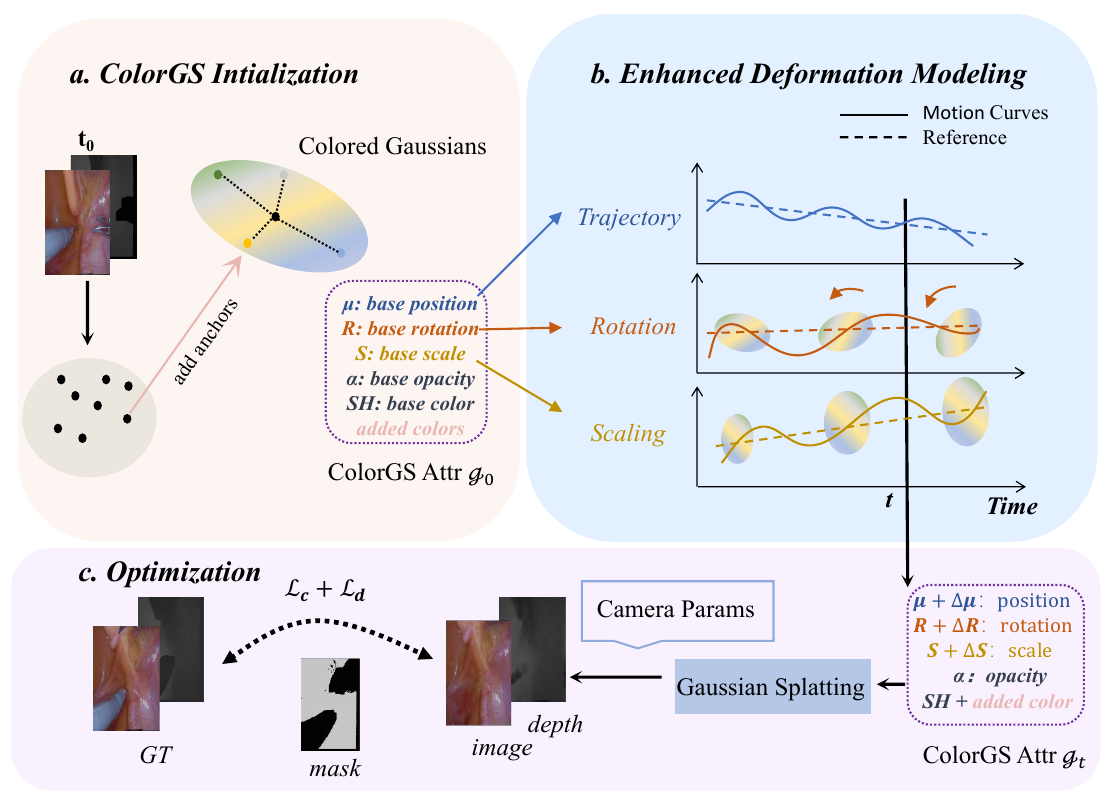}
\caption{Pipeline of ColorGS, composed of (a) ColorGS Initialization, (b) Enhanced Deformational Modeling, and (c) Optimization using color and depth loss functions. }\label{fig1}
\end{figure}

We propose ColorGS, a novel deformable surgical scene reconstruction paradigm that combines Colored Gaussians and EDM to achieve high-quality and efficient 3D dynamic tissue reconstruction.
As shown in Fig. \ref{fig1}, our method begins by using the corresponding camera parameters to obtain the point cloud of the first frame, which is then used to initialize the Gaussian primitives (Sec. \ref{3DGS}).
During the optimization, spatially adaptive anchor color aggregation is employed to enhance the color expressiveness of the Gaussian primitives (Sec. \ref{sec:1}).
Besides, a combination of time-aware basic functions and time-independent global deformation parameters is utilized to model the dynamic changes of the Gaussian primitives' properties (Sec. \ref{sec:2}).
Finally, the whole paradigm is trained using two loss functions that compare the ground-truth (GT) images with the rendered RGB/depth images (Sec. \ref{sec:optimization}).

%

\subsection{Preliminaries of 3D Gaussian Splatting} \label{3DGS}
3DGS \cite{3DGS} introduces anisotropic 3D Gaussian parameters and a tile-based rasterizer to achieve high-quality and real-time 3D scene reconstruction.
Each Gaussian is parameterized by its center position $\mu$, covariance matrix $\Sigma$, opacity $o$, and spherical harmonic (SH) coefficients.
For an arbitrary coordinate $x$, the shape of the 3D Gaussian on $x$ is described as:
\begin{equation}\label{eq:1}
G(x) = exp(-\frac{1}{2}(x-\mu)^T\Sigma^{-1}(x-\mu)),
\end{equation}
\begin{equation}
\Sigma=RSS^TR^T
\end{equation}
where the covariance matrix $\Sigma$ is decomposed into a rotation matrix $R$ and a diagonal scaling matrix $S$, allowing independent control over orientation and extent.  
All 3D Gaussians are projected onto the 2D image plane along the rays, enabling efficient rendering through fast $\alpha$-blending.
The positions $\mu^{2D}$ and the covariance matrices $\Sigma^{2D}$ of the projected 2D Gaussians can be analytically computed in the pixel coordinate system using the camera intrinsic and extrinsic parameters.
To achieve view-independent rendering, the color $\hat{C}(x)$ and the depth $\hat{D}(x)$ of a center pixel $x$ can be rendered by the function:
\begin{equation}
\hat{C}(x)=\sum_{i=1}^nc_i\alpha_i\prod_{j=1}^{i-1}(1-\alpha_j), \hat{D}(x)=\sum_{i=1}^nd_i\alpha_i\prod_{j=1}^{i-1}(1-\alpha_j)
\end{equation}
where $\alpha_i$ is given by evaluating a 2D covariance matrix $\Sigma^{2D}$ multiplied by the opacity $o_i$, and $c_i$ is the color computed from the SH coefficients of the $i$-th Gaussian, which is fixed for each Gaussian after the optimization process.



\subsection{Colored Gaussian Primitives} \label{sec:1}
In surgical scenes, variable lighting conditions and subtle color transitions often challenge the color expression capability of the original 3D Gaussians with fixed colors, leading to suboptimal reconstruction performance.
We notice that adjacent tissues in surgical scenes typically have similar appearances, and identifying differences between these regions requires a careful comparison of neighboring pixels.
This observation highlights the importance of incorporating spatial properties into the color modeling process for surgical scenes.
Inspired by ~\cite{NegGS,3D-HGS,superGaussian}, we introduce a group of color anchors for each Gaussian to provide spatial-aware hints for optimizing its color representation, thereby enabling the capture of subtle color variations between similar tissues. 
These enhanced representations are referred to as Colored Gaussians.

Specifically, we introduce $k$ dynamic anchors $A_i = (A_i^x, A_i^y)$ with color parameters $c_i$ on the rendering plane for each Gaussian primitive, where $i = 0, 1, 2, . . . , k-1$ corresponds to the index of anchors. 
Then we use an exponential decay function to measure the contribution of each anchor to the color at the intersection point.
When the ray intersects with the Gaussian primitive and generates the intersection point $p = (u, v)$ on the rendering plane, the decay rate depends on the distance between $p$ and the anchor coordinate $A_i$, which can be defined as:
\begin{equation}\label{eq:5}
F_{A_i}(p) = e^{-\lambda _e||p-A_i||^2}
\end{equation}
where $\lambda _e$ is used to control the rate of change. We set $\lambda _e$ = 0.1 and $k$ = 4 by default. 
The added color produced by these anchors can be computed as:
\begin{equation}\label{eq:4}
F_c(p) = \sum _{i=0}^{k-1}F_{A_i}(p)c_i
\end{equation}
As a result, the color function for each Gaussian primitive is represented as: 
\begin{equation}\label{eq:3}
c(p, d) = SH(d) + F_c(p)
\end{equation}
where $d$ represents the direction of the ray, $SH(d)$ represents the spherical harmonics of direction $d$. 

By introducing these anchors, each individual Gaussian can reflect different color variations at the intersection points, better fitting the details.
Besides, since these anchors are defined on the rendering plane, the additional colors associated with the Colored Gaussians are inherently view-dependent, further enhancing spatially aware color modeling.

\subsection{Enhanced Deformation Modeling} \label{sec:2}
Modeling Gaussian motion through linear combinations of time-aware basis functions proves to be an efficient strategy~\cite{Gaussian-flow,spacetime,Deform3DGS}, particularly for real-time surgical scene reconstruction.
Among the available basis functions, Gaussian functions are preferred over the commonly used Fourier and polynomial basis functions.
This preference arises from their ability to offer localized influence to preserve details without disrupting global motion.
The Gaussian basis functions can be defined as:
\begin{equation}
\tilde{b}(t;\theta_j^x,\sigma_j^x)=exp(-\frac{(t-\theta)^2}{2\sigma^2})    
\end{equation}
where $t$ represents time, and $\theta$ and $\sigma$ correspond to the learnable center and variance, respectively.

However, due to the localized nature of Gaussian functions, modeling consistent long-term motion often requires a larger number of basis functions and involves more complex parameter optimization.
Besides, when specific motion patterns arise, using linear combinations of Gaussian functions to accurately model them can be challenging. For instance, uniform motion requires a substantial number of Gaussian functions for proper fitting.
To solve it, we propose EDM, which decouples motion into local dynamics represented by linear combinations of time-aware Gaussian functions and global motion trends modeled by time-independent global motion parameters.
Specifically, the deformation parameters include the center position $\mu$, the rotation matrix $R$, and the scaling matrix $S$.
Taking the center position change of the Gaussian in the $x$-direction as an example, the position at any time $t$ can be expressed as:
\begin{equation}\label{eq:6}
\psi^x(t, \Theta^x) = \sum _{j=0}^{B-1}\omega_{j}^x\tilde{b}(t;\theta_j^x,\sigma_j^x)+\delta_x
\end{equation}
where $B = 17$ denotes the total number of basis functions.
EDM decouples the global motion trend, which is challenging to represent with Gaussian functions, allowing for more flexible modeling of diverse motion patterns and enhancing the overall consistency and smoothness of the deformation.

\subsection{Optimization}\label{sec:optimization}
We optimize the whole paradigm by minimizing the discrepancy between the rendered outputs and the GT images.
The whole training loss functions $\mathcal L_{total}$ can be defined as:
\begin{equation}\label{eq:7}
\mathcal L_{total} = \|M\odot(\hat{C}-C)\|+ \|M\odot(\hat{D}-D)\|
\end{equation}
where $\hat{C}$ and $\hat{D}$ denote the rendered RGB and depth images,  $C$ and $D$ represent the GT RGB and depth images, and $M$ is the tissue mask.

\section{Exeperiment}
\subsection{Experiment Setting}

\subsubsection{Dataset and Evaluation.} Following \cite{Deform3DGS}, we evaluate the proposed method on two datasets: EndoNeRF~\cite{EndoNeRF} and StereoMIS~\cite{StereoMIS}. EndoNeRF contains two cases of in-vivo prostatectomy data captured from stereo cameras at a single viewpoint, encompassing challenging scenes with non-rigid deformation and tool occlusion. The video clips in StereoMIS are captured from in-vivo porcine subjects and present additional challenges, including diverse anatomical structures and significant tissue deformations. All scenes of EndoNeRF and three clips from videos P2 and P3 in StereoMIS are used for performance evaluation. 
For each scene, the frames are divided into training and testing sets with a ratio of 7:1. To quantify the performance, we use PSNR, SSIM, and LPIPS as the metrics.

\subsubsection{Implementation Details.} For each scene, the video duration is normalized into $[0, 1]$. The training process spans 3000 iterations, starting with an initial learning rate of $1.6\times10^{-3}$. The densification on the Gaussian points number is frozen during the first 600 iterations for stable training. All the experiments are conducted with an NVIDIA RTX 3060 GPU.

\subsection{Comparison with State-of-the-art Methods}
\begin{figure}
\includegraphics[width=\textwidth]{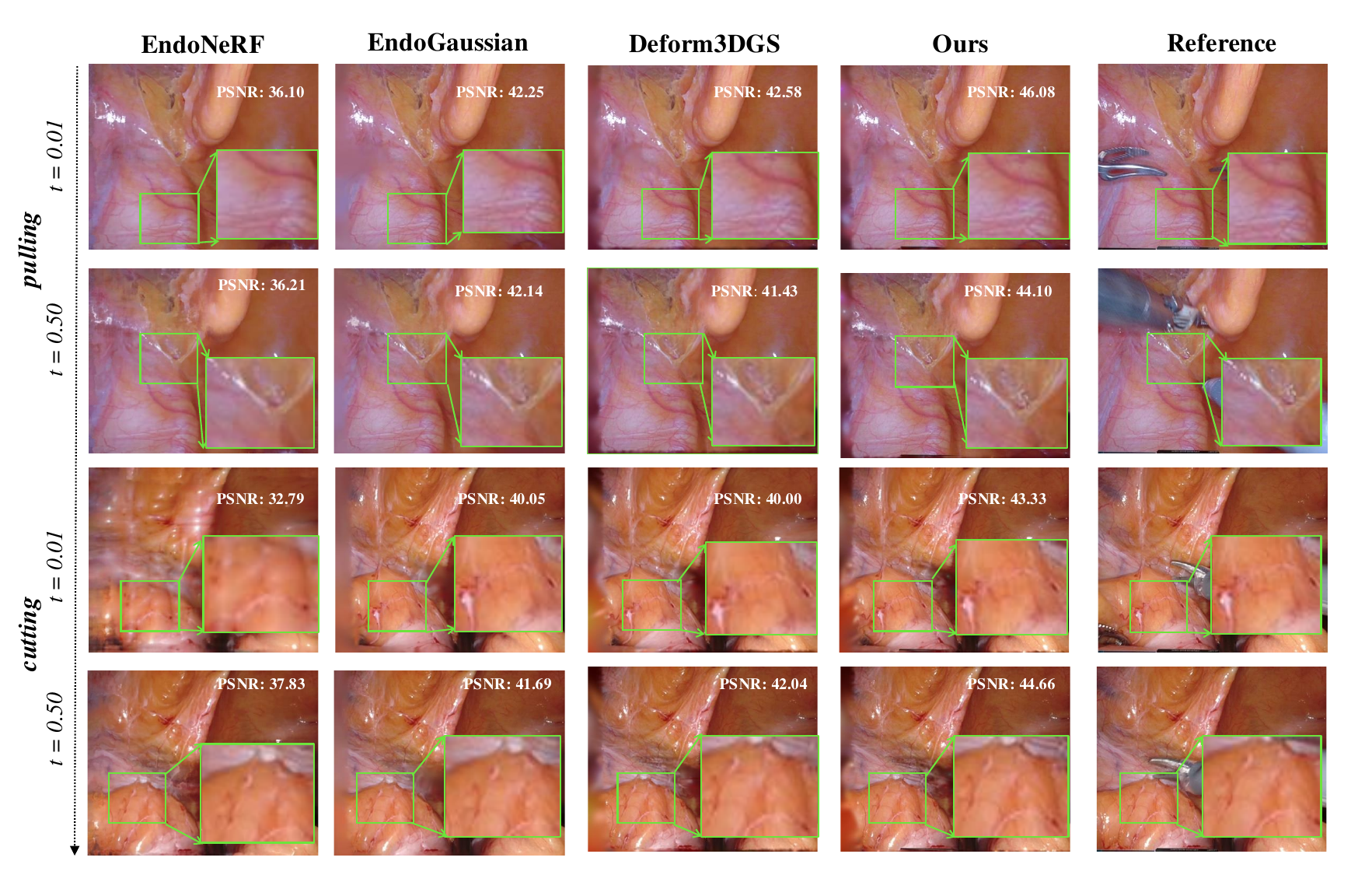}
\caption{Illustration of the rendered images of baselines and ours.} \label{fig2}
\end{figure}

Our proposed framework is compared with EndoNeRF and two recent 3DGS-based methods, i.e., EndoGaussian~\cite{EndoGaussian} and Deform3DGS~\cite{Deform3DGS}. 
EndoNeRF suffers from long training times and poor performance, severely limiting its practical use in surgery. EndoGaussian introduces Gaussian functions into dynamic surgical scene reconstruction and models dynamic Gaussians by decomposing the feature plane. Building on this, Deform3DGS further proposes the use of basis functions instead of traditional MLP networks for more efficient reconstruction. However, traditional 3D Gaussians struggle with limited color expressiveness, making it difficult to capture the similar appearance and subtle color variations of surgical tissues. Moreover, due to the localized nature of basis functions, capturing consistent global motion trends remains challenging. Our method overcomes these limitations, significantly improving both color accuracy and global motion consistency, resulting in higher-quality surgical scene reconstruction.
As shown in Table \ref{tab1} and Fig. \ref{fig2}, our method outperforms recent state-of-the-art surgical scene reconstruction results by a large margin. 
Specifically, on the EndoNeRF dataset, we achieve a remarkable PSNR improvement of 1.50 dB over the second-best method, Deform3DGS, while also improving SSIM and LPIPS.
%
%
%
%

\begin{table}
\caption{Quantitative evaluation on endoscopic scene reconstruction. The best and suboptimal results are shown in \textbf{bold} and \underline{underlined} respectively.}\label{tab1}
\begin{tabular}{>{\centering\arraybackslash}p{3cm}|>{\centering\arraybackslash}p{3cm}|>{\centering\arraybackslash}p{2cm}>{\centering\arraybackslash}p{2cm}>{\centering\arraybackslash}p{2cm}}
\toprule
Dataset &  Method & PSNR$\uparrow$ & SSIM(\%)$\uparrow$ &  LPIPS$\downarrow $\\
\midrule
\multirow{4}{*}{EndoNeRF} & EndoNeRF & 35.92 & 94.18 & 0.06\\ 
& EndoGaussian & 37.86 & 96.09 & \underline{0.04}\\ 
& Deform3DGS & \underline{38.35} & 96.39 & 0.05 \\ 
& ours & \textbf{39.85} & \textbf{97.25}  & \textbf{0.03}\\ 
\midrule
\multirow{4}{*}{StereoMIS} & EndoNeRF & 28.86 & 74.15 & 0.27 \\ 
& EndoGaussian & 30.39 & 83.75 & \underline{0.21}\\
& Deform3DGS & \underline{30.68} & \underline{84.74} & 0.23\\
& Ours & \textbf{32.64} & \textbf{89.64}  & \textbf{0.14}  \\
\bottomrule
\end{tabular}
\end{table}

\begin{table}
\caption{Ablation study of the designed components on EndoNeRF.}\label{tab2}
\begin{tabular}{>{\centering\arraybackslash}p{3cm}|>{\centering\arraybackslash}p{3cm}|>{\centering\arraybackslash}p{2cm}>{\centering\arraybackslash}p{2cm}>{\centering\arraybackslash}p{2cm}}
\hline
Component & Method & PSNR$\uparrow$ & SSIM(\%)$\uparrow$ &  LPIPS$\downarrow $ \\
\hline
\multirow{4}{*}{Gaussian Splatting} & 2DGS & 36.16 & 95.99 & 0.04 \\ 
& 3DGS & 38.55 & 97.08 &  0.03 \\
& SuperGS & 39.31 & 97.13 & 0.03 \\
& Ours & \textbf{39.85} & \textbf{97.25}  & \textbf{0.03}\\
\hline
\multirow{3}{*}{Gaussian tracking} 
& FPS & 38.92 & 96.69 & 0.04 \\ 
& GS & 39.54 & 97.13 & 0.03 \\ 
& Ours & \textbf{39.85} & \textbf{97.25}  & \textbf{0.03}\\
\hline
\end{tabular}
\end{table}

\subsection{Quantitative Evaluation of Key Components}
\subsubsection{The Effect of Colored Gaussians.} We conduct experiments using different 3D representations, including the original 3DGS and 2D Gaussian Splatting (2DGS)~\cite{2DGS}, to evaluate the color expression capabilities of our Colored Gaussian. 
As shown in Table \ref{tab2}, both 3DGS and 2DGS exhibit inferior performance compared to the proposed Colored Gaussian. 
Additionally, our method also outperforms SuperGaussian \cite{superGaussian}, which is built upon 2DGS, in terms of rendering quality.
This demonstrates that the proposed method can effectively enhance the overall quality of the complex surgical scene reconstruction.


\subsubsection{The Effect of Enhanced Deformation Modeling.} 
To investigate the effectiveness of EDM, we compare it with existing deformation modeling techniques on the EndoNeRF dataset. 
Specifically, we replace EDM with alternative methods, including the combination of Fourier and Polynomial series (FPS) and the Gaussian functions without global parameters (GS). 
As shown in Table \ref{tab2}, EDM achieves the best performance in representing deformations.

\section{Conclusion}
In this paper, we propose a high-quality dynamic modeling method for surgical scenes, called ColorGS. To enhance the expressive power of Gaussian representation, we develop Colored Gaussians, which introduce spatially varying colors to significantly improve the flexibility and precision of color representation for individual Gaussians. Additionally, to capture global deformation trends, we propose the Enhanced Deformation Model, which leverages basis functions and innovatively incorporates time-independent global motion parameters to ensure both global consistency and local adaptability. By seamlessly combining these two quality-enhancement strategies, we achieve high-quality surgical scene reconstruction with enhanced color details and deformation accuracy.

\bibliographystyle{splncs04}
\bibliography{mybibliography}
\end{document}